\documentclass[letterpaper, 10 pt, conference]{ieeeconf}  

\IEEEoverridecommandlockouts                              

\usepackage[font=footnotesize, tableposition=top]{caption}

\usepackage{times}
\usepackage{epsfig}
\usepackage{graphicx}
\usepackage{amsmath}
\usepackage{amssymb}
\usepackage{color}
\usepackage{multirow}
\usepackage{subcaption}
\usepackage{transparent}
\graphicspath{{images/}}

\usepackage{algorithm}
\usepackage[noend]{algpseudocode}

\usepackage{color}
\definecolor{darkred}{rgb}{0.5, 0.0, 0.0}
\definecolor{darkblue}{rgb}{0.0, 0.0, 0.5}
\definecolor{darkgreen}{rgb}{0.0, 0.5, 0.0}
\newcommand{\hide}[1]{}

\renewcommand{\baselinestretch}{0.92}

\title{\LARGE \bf
Action Image Representation: Learning Scalable Deep Grasping Policies with Zero Real World Data
}

\author{Mohi Khansari$^{1}$, Daniel Kappler$^{1}$, Jianlan Luo$^{2}$, Jeff Bingham$^{1}$, Mrinal Kalakrishnan$^{1}$
\thanks{$^{1}$X, The Moonshot Factory,
        {\tt\small \{khansari, kappler, jeffbingham, kalakris\}@google.com},
        }%
\thanks{$^{2}$University of California, Berkeley (the work was done as an intern at X.),
        {\tt\small jianlanluo@eecs.berkeley.edu}}%
}

\begin{document}

\maketitle
\thispagestyle{empty}
\pagestyle{empty}

\begin{abstract}
This paper introduces \textit{Action Image}, a new grasp proposal representation that allows learning an end-to-end deep-grasping policy. Our model achieves $84\%$ grasp success on $172$ real world objects while being trained only in simulation on $48$ objects with just naive domain randomization.
Similar to computer vision problems, such as object detection, \textit{Action Image} builds on the idea that object features are invariant to translation in image space. Therefore, grasp quality is invariant when evaluating the object-gripper relationship; a successful grasp for an object depends on its local context, but is independent of the surrounding environment.
\textit{Action Image} represents a grasp proposal as an image and uses a deep convolutional network to infer grasp quality. 
We show that by using an \textit{Action Image} representation, trained networks are able to extract local, salient features of grasping tasks that generalize across different objects and environments.
We show that this representation works on a variety of inputs, including color images (RGB), depth images (D), and combined color-depth (RGB-D). 
Our experimental results demonstrate that networks utilizing an \textit{Action Image} representation exhibit strong domain transfer between training on simulated data and inference on real-world sensor streams.
Finally, our experiments show that a network trained with \textit{Action Image} improves grasp success ($84\%$ vs. $53\%$) over a baseline model with the same structure, but using actions encoded as vectors.
\end{abstract}

\section{Introduction}
Grasping is a fundamental task that is a prerequisite for robots to manipulate the environment and interact with the world.
Common household tasks, such as tidying a room, setting a table, or preparing a meal all require grasping as a foundational capability.
Due to its importance for manipulation, nearly every facet of grasping has been studied in the field of robotics.
In this paper, we build on previous work on inferring grasp poses through models and propose a new representation, \textit{Action Image}, to predict successful grasps from perceptual observations.

\begin{figure}[h]   
  \begin{center}
  \includegraphics[width=\linewidth]{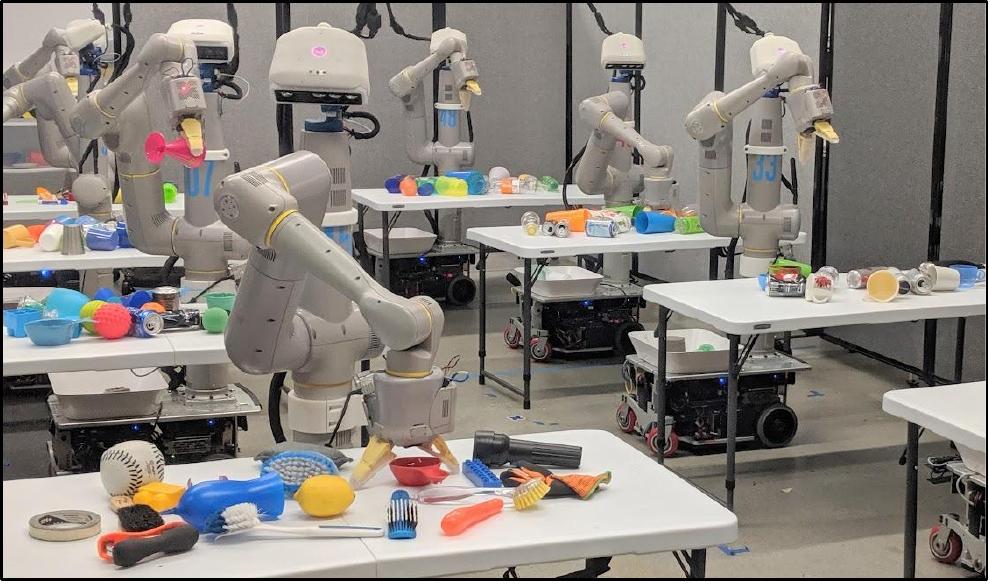} 
  \end{center}
  \caption{\footnotesize{Real-world grasp success evaluation of policies trained only in simulation using a cluster of 8 mobile manipulators.}}
  \label{fig:af_experiment}   
\end{figure}

A common practice for data-driven grasping is a two-stage method of first using perception to detect objects and/or estimate their poses in a scene, and then synthesizing a feasible grasp using a planner or a learned policy in the robot's task-space. 
We refer to these types of methods as perception-dependent grasp synthesis approaches, see \cite{bohg2013data} for an extensive literature review. 
More recently, there has been a shift towards end-to-end grasping, inferring grasps directly from sensory information such as images and/or depth.

End-to-end approaches compare favorably to two-stage perception-dependent methods, since they are jointly optimized and the overall performance is not bound by the performance of each individual stage.
This improvement is in large part due to the difficulty in recovering from errors. 
Multistage methods benefit from abstractions that allow solving a collection of easier sub-problems, but the downside is each stage increases the possibility of irrecoverable failure.
As an example, if the perception stage poorly estimates the pose of an object, then all subsequent grasp synthesis will accumulate this error.
In contrast, end-to-end grasping policies learn implicit object-centric representations where object-detection and pose estimation become a fused part of the network and multiplicative errors are avoided.
The cost of a combined representation is typically the need for massive datasets to produce quality results and the increased complexity of optimization during training.
While these large datasets are possible to obtain in simulation, it is often infeasible to obtain the same scale of data in the real-world.
A further concern is that most end-to-end methods do not generalize well to data distribution shifts, which is unfortunately very common when transferring from synthetic to real-world data.

End-to-end approaches for grasping are roughly formulated to solve a one-to-many problem. 
This is because grasping is naturally multimodal; given a single observation there exist many grasp poses that will lead to a successful grasp. 
There are two predominant end-to-end approaches for tackling this: 
1) Stochastic \textit{actor networks} that produce a multimodal distribution between a sensory observation and possible grasps \cite{mengyuan19}.
2) \textit{Critic networks} that produce a map between sensory observation with grasp proposal and probability of success \cite{levine2018learning}.
In this paper we choose to focus on a \textit{critic network} approach, building on the recently demonstrated effectiveness of its use in grasping by \cite{levine2018learning}. 

We hypothesize that by representing grasps (actions) with a representation that is compatible with the intrinsic assumptions of the network predictions of grasp success, \textit{it is possible to learn grasps for novel objects in the real world without the necessity of any real world training data.} 
We argue that a meaningful representation for grasping must be:
1) Applicable to common sensory modalities accessible to robots; namely color images, depth images, and their combination,
2) Domain invariant, such that it can seamlessly work in either simulated or real environments; in other words, the representation of grasp quality should be independent of whether the sensory input is synthetic or real,
3) Generalizable to new objects, and
4) Invariant to translations in sensor-space; grasp experience should be efficiently transferred between locations of the task-space. 

We test our hypothesis by proposing a novel representation, called \textit{Action Image}, that contains the above characteristics through embedding a grasp proposal as rendered features and included as channels of an \textit{image}. 
This representation allows us to build a deep-convolutional grasp policy that innately is invariant to translation in image space, assuming both the object and grasp is transformed in the same manner; information about the grasp hypothesis naturally fits the implicit structural assumption imposed by convolutions.

Furthermore, this paper provides experimental validation of our hypothesis that \textit{Action Image} is a meaningful representation that improves grasp success prediction and facilitates domain transfer.
We first demonstrate performance gains and compare the efficiency of the method across increasing amounts of simulated data. 
Then, we provide strong evidence of domain invariance with transfer to real-world scenarios, showing that \textit{Action Image} networks trained with synthetic data generalize to real world objects with only slightly degraded performance

In summary, the contributions of this paper are as follows:
\begin{itemize}
\item A new representation, called \textit{Action Image}, that makes grasp prediction invariant to translation in image space.
\item Demonstrating how an \textit{Action Image} representation is easily extensible to different input modalities, e.g. RGB, D, and RGB-D images.
\item Experimentally verifying the data efficiency and the domain invariant performance of our method.
\end{itemize}

\section{Related Work}\label{sec:related_work}
Grasping is a building block for manipulation and has long been explored by the robotics community~\cite{bicchi2000robotic,bohg2013data}.
Early approaches to grasping used models of the environment coupled with grasp planning based on force closure metrics~\cite{bicchi1995closure} or grasping-specific simulations~\cite{miller2004graspit} to compute optimal grasps. 
In order to execute these predicted grasps in the real world, the 3D object models have to be identified, located and their pose estimated~\cite{goldfeder2011data,hernandez2016team,kehoe2013cloud,ciocarlie2014towards}. 

Recently, data-driven methods have become popular~\cite{bohg2013data}, due to their flexibility in specifying sensory input and ease of incorporating data from experience through real robot grasp executions \cite{herzog2014learning}.
Some of the first work to apply deep learning to the problem of grasp detection used manually annotated datasets of images and corresponding grasps. 
These approaches applied two-stage learning pipelines similar to those used in object detection and classification in computer vision~\cite{lenz2015deep, nguyen2016detecting}.
Since manual annotation can be time-consuming, simulation approaches have been exploited to generate real-world like sensory data~\cite{kappler2015leveraging,mahler2017dex, varley2015generating, veres2017modeling}.
For real world data collection researchers have since moved towards self-supervised data collection, wherein a robot labels its own grasps based on heuristics~\cite{pinto2016supersizing}.
Pushing self-supervised learning to millions of trials has allowed learning of hand-eye coordination for grasping~\cite{levine2018learning}, as well as applying a variant of Q-learning~\cite{pmlr-v87-kalashnikov18a} to grasp unknown objects in a cluttered environment using a single color camera. 
These aforementioned data-driven approaches are related to our work, since they train a critic network that accepts images and actions as inputs and output a metric of grasp success.
In our work we also use the cross-entropy method (CEM)~\cite{rubinstein1999cross} to find optimal actions from a critic network. 
Our work is differentiated from previous efforts in a few ways: 
(1) we represent actions directly in image space, which simplifies the network and allows learning from just simulated data, and 
(2) our network is easily extensible to use depth data.

With the rise of data-driven techniques considerable attention has been given to leveraging simulation to reduce the need for real-world robot data, which is tedious and time-consuming to collect.
Researchers have used domain randomization~\cite{tobin2017domain,sadeghi2018sim2real,james2017transferring} or trained pixel-to-pixel networks, converting either from simulated images to real images~\cite{bousmalis2018using}, or vice versa~\cite{james2019sim}.
Compared to \cite{james2019sim}, our network is trained supervised, significantly smaller and faster to train, as it does not rely on another network to transfer images from one domain to another.

Work in this area has also demonstrated that using depth data~\cite{mahler2017dex} (as opposed to RGB images alone) results in better performance when transferring policies trained in simulation directly to real robots.
This observation also motivated our experimental design to test improvements in performance using depth data over RGB data alone. 
Compared to \cite{mahler2017dex}, our representation is applicable to both RGB and depth, and allows easier extension to higher action dimensions such as 6D grasping  or the gripper angle.

Fully convolutional networks have also been leveraged to perform grasp detection in image space by producing several output heatmaps indicating grasp success discretized over top-down gripper orientation~\cite{zeng2018robotic}.
While this approach shares some of the advantages of our grasp representation, such as translation-invariance, our approach naturally extends to higher degrees of freedom without the need to further discretize the output heatmap space.

\begin{figure}[ht]   
  \begin{center}
  \includegraphics[width=\linewidth]{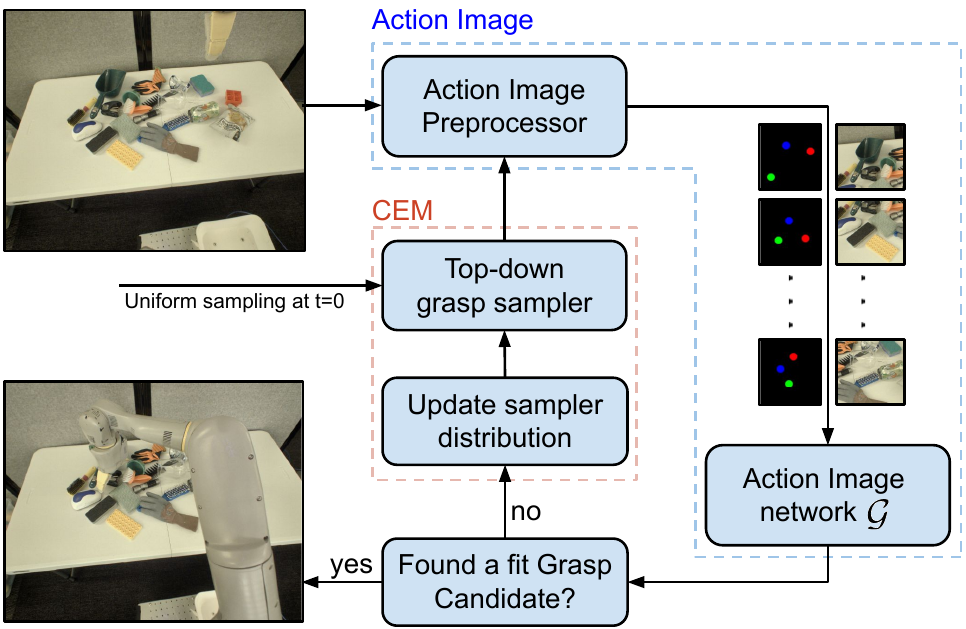} 
  \end{center}
  \caption{\footnotesize{The procedure for performing end-to-end grasping using an \textit{Action Image} representation. Our mobile manipulator first approaches the workspace and then grasp candidates are uniformly sampled from the workspace. Next, a preprocessor generates pairs of sensed-image and action-image data corresponding to crops centered around the grasp proposals. Then, an \textit{Action Image} network evaluates the candidate pairs and if a grasp proposal exceeds the desired probability of success the grasp will be executed by the robot. Otherwise, the grasp candidate distribution is updated with the top $M\%$ of elite proposals and the process is repeated.}
  \vspace{-0.4cm}
  }
  \label{fig:architecture}   
\end{figure}

\section{Method} \label{sec:method}

\begin{figure}[h]   
  \centering
    \includegraphics[width=0.77\linewidth]{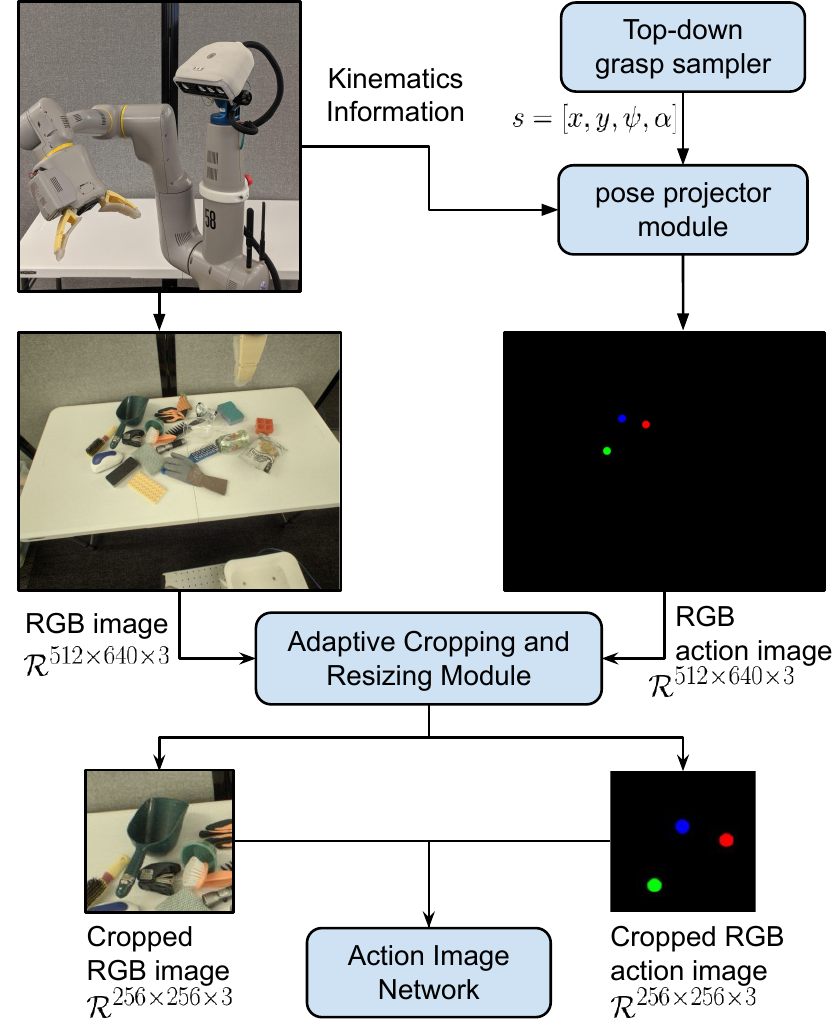}
  \caption{\footnotesize{The description of the \textit{Action Image} preprocessor. Given a sample pose, the 3D position of the gripper features are calculated and projected into camera-space producing the \textit{Action Image} representation. The number of these features  determines the expressiveness of the representation and at least 3 points are required to estimate the pose. The values of the projected points can carry additional information, such as depth in camera frame. Next, the resulting \textit{Action Image} and sensor-image data are adaptively cropped to a $128 \times 128 \times 3$ square centered at the proposed grasp candidate and then are resized to $256 \times 256 \times 3$.}
  \vspace{-0.2cm}
  }
  \label{fig:preprocessor}   
\end{figure}

\begin{figure}[t]   
  \begin{center}
  \vspace{-0.2cm}
  \includegraphics[width=0.8\linewidth]{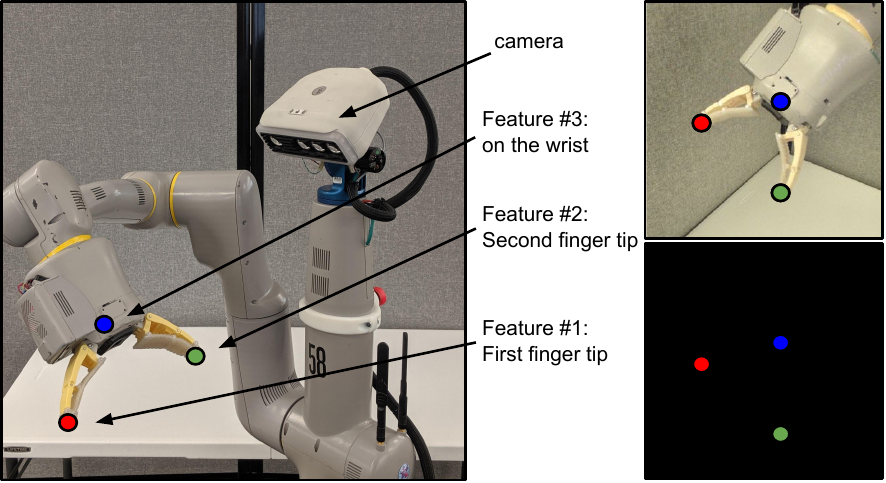} 
  \end{center}
  \vspace{-0.1cm}
  \caption{\footnotesize{\textbf{Left:} We represent a gripper by 3 feature points: two features are placed on the gripper finger tips and the third one is placed on the point where the fingers are attached to the wrist. \textbf{Right:} These feature points can be easily projected into camera-space through known kinematic transformations available from the robot. The generated \textit{Action Image} is shown in the bottom-right image.
  \vspace{-0.4cm}
  }
  }
  \label{fig:representation_kinematics}   
\end{figure}

\begin{figure}[ht]
    \centering
    \begin{subfigure}[b]{0.3\linewidth}
        \includegraphics[width=\linewidth]{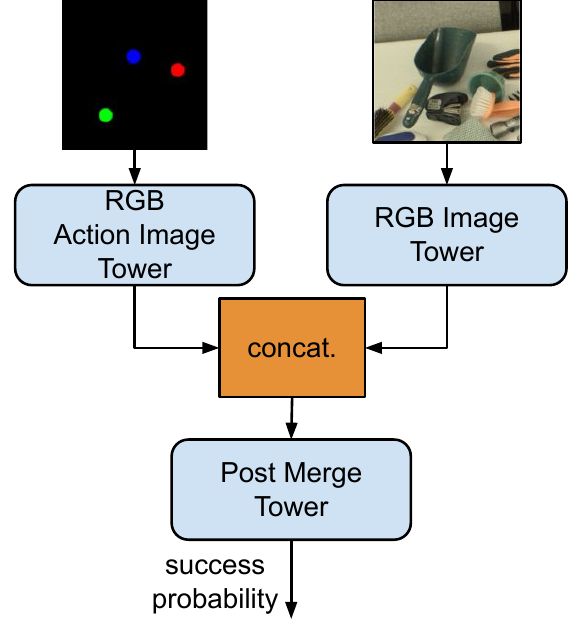}
        \caption{\footnotesize{Action RGB image net.}
        }
        \label{fig:action_image_net}
    \end{subfigure}\hspace{5mm}
    \begin{subfigure}[b]{0.3\linewidth}
        \includegraphics[width=\linewidth]{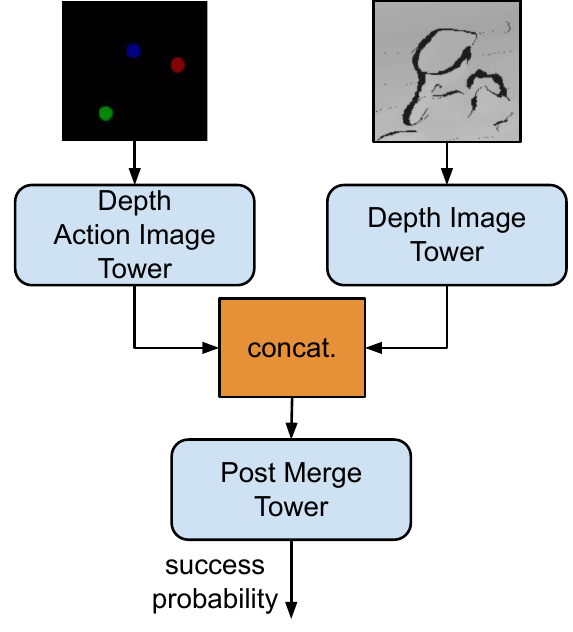}
        \caption{\footnotesize{Action depth image net.}}
        \label{fig:action_depth_net}
    \end{subfigure}
    \begin{subfigure}[b]{0.6\linewidth}
        \includegraphics[width=\linewidth]{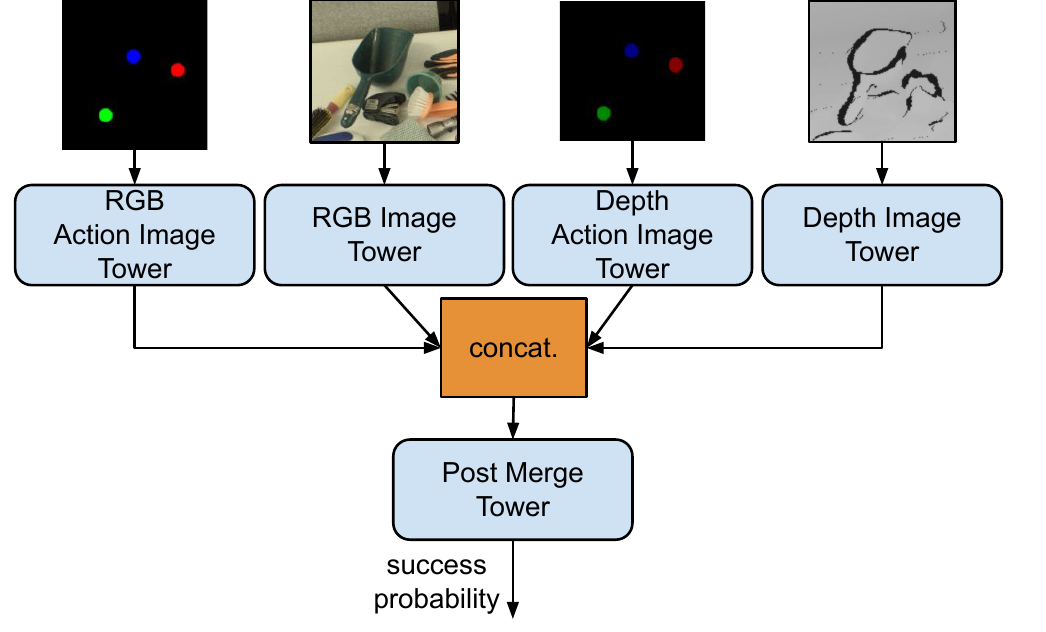}
        \caption{\footnotesize{Action RGB-D image net.}}
        \label{fig:action_image_depth_net}
    \end{subfigure}
    \begin{subfigure}[b]{0.3\linewidth}
        \includegraphics[width=\linewidth]{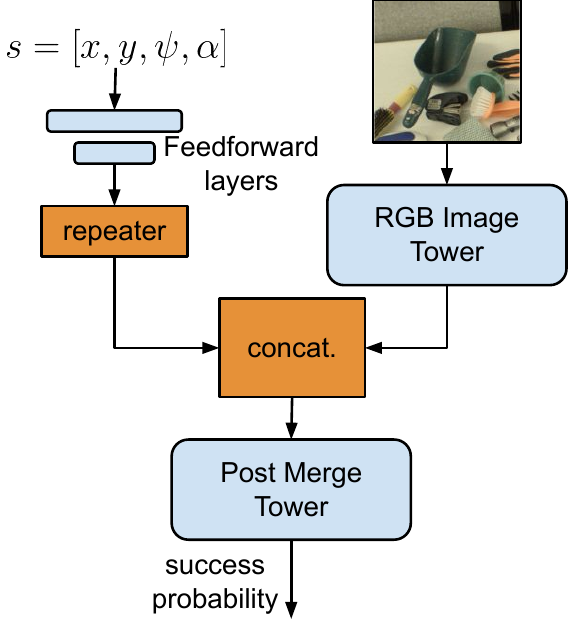}
        \caption{\footnotesize{Baseline: Action vector net.}}
        \label{fig:action_vector_net}
    \end{subfigure}
    \vspace{-0.2cm}
    \caption{\footnotesize{Representations for the \textit{Action Image} networks used in this paper were (a) action RGB (b) action D and (c) action RGB-D. All networks were composed solely of convolutional layers. These networks were compared to a baseline (d) action vector network where the grasp candidates were fed to the network as a 4D vector of floats $[x, y, \psi, \alpha]$.}
    \vspace{-0.3cm}
    }
    \label{fig:nets}
\end{figure}

\begin{figure}[ht]
    \centering
    \vspace{-0.2cm}
    \begin{subfigure}[b]{\linewidth}
        \includegraphics[width=0.9\linewidth, trim=0 0.57cm 0 0, clip]{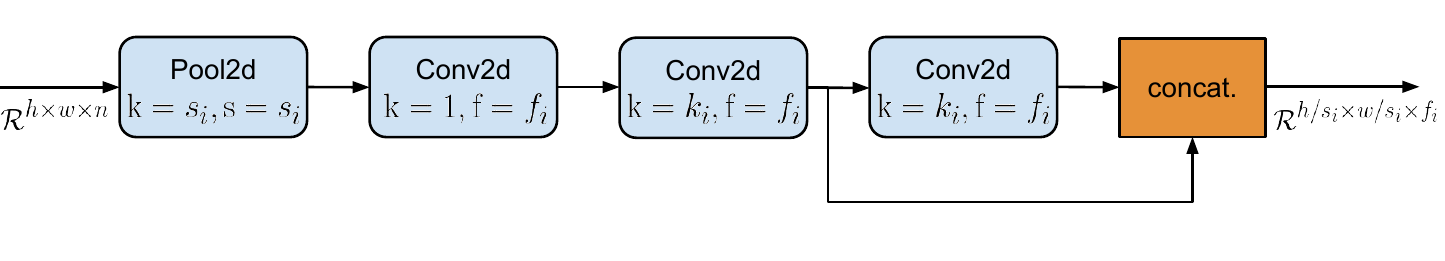} 
        \caption{\footnotesize Resnet Layer. \vspace{-0.1cm}}
        \label{fig:resnet}
    \end{subfigure}
    \begin{subfigure}[b]{\linewidth}
        \centering
        \includegraphics[width=0.8\linewidth, trim=0 1.3cm 0 0, clip]{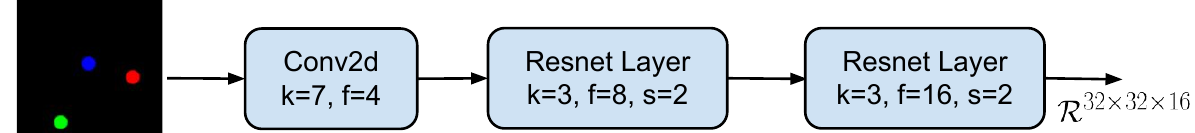}
        \caption{\footnotesize RGB action image tower. \vspace{-0.1cm}}
        \label{fig:tower_action_image_rgb}
    \end{subfigure}
    \begin{subfigure}[b]{\linewidth}
        \centering
        \includegraphics[width=0.8\linewidth, trim=0 1.3cm 0 0, clip]{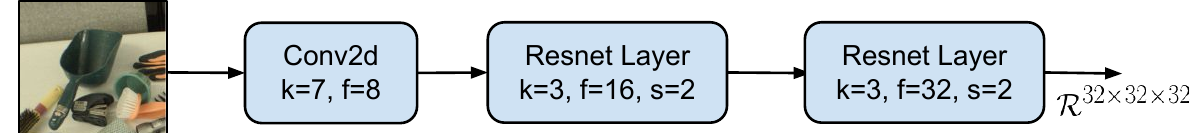}
        \caption{\footnotesize RGB image tower. \vspace{-0.1cm}}
        \label{fig:tower_image_rgb}
    \end{subfigure}
    \begin{subfigure}[b]{\linewidth}
        \centering
        \includegraphics[width=0.8\linewidth, trim=0 1.3cm 0 0, clip]{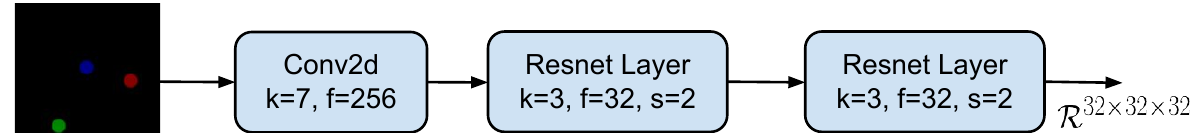}
        \caption{\footnotesize Depth action image tower. \vspace{-0.1cm}}
        \label{fig:tower_action_image_d}
    \end{subfigure}
    \begin{subfigure}[b]{\linewidth}
        \centering
        \includegraphics[width=0.8\linewidth, trim=0 1.3cm 0 0, clip]{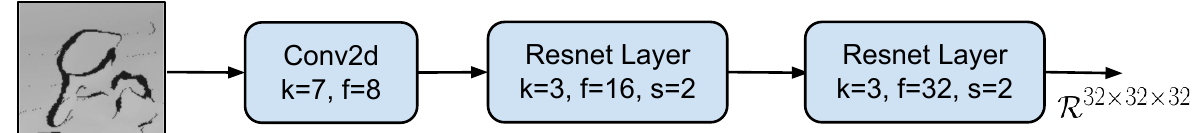}
        \caption{\footnotesize Depth image tower. \vspace{-0.05cm}}
        \label{fig:tower_image_d}
    \end{subfigure}
    \begin{subfigure}[b]{\linewidth}
        \centering
        \includegraphics[width=0.8\linewidth, trim=0 1.3cm 0 0, clip]{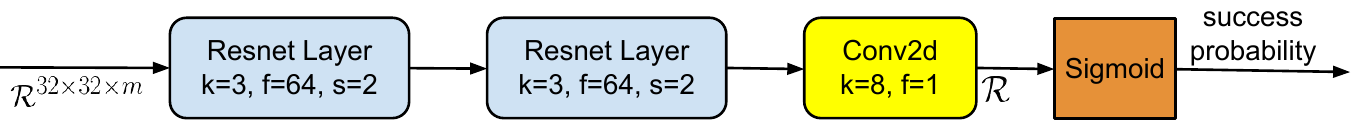}
        \caption{\footnotesize Post merge tower. \vspace{-0.1cm}}
        \label{fig:tower_post_merge}
    \end{subfigure}
    \vspace{-0.4cm}
    \caption{
    \footnotesize{Details that illustrate the sub-components of the different action image networks.}
    \vspace{-0.2cm}
    }
    \label{fig:towers}
\end{figure}

The grasping problem was formulated as an iterative optimization procedure using a supervised critic network approach, similar to \cite{levine2018learning}. 
A learned critic network $\mathcal{G}$ was used to evaluate the likelihood of grasp success for the inputs of sensor observations and action representations for table-top grasps.
The Cross Entropy Method (CEM) was used in conjunction to refine the best grasp candidate by iteratively sampling in an action space~\cite{rubinstein1999cross}. 

The grasp prediction process is laid out in Fig.~\ref{fig:architecture}.
The initial sample distribution at $t=0$ is a design parameter and without lack of generality this implementation used a uniform distribution over a predefined robot workspace for simplicity.
All following samples were fed through the \textit{Action Image} Preprocessor (Section~\ref{sec:representation}) that generated the network input representation based on the grasp proposals.
Next, the critic network $\mathcal{G}$ evaluated the probability of success $\mathcal{P}$ for all grasp candidates.
When the grasp with the highest probability exceeded a threshold $\mathcal{P} \geq \mathcal{P}_{thr}$ that grasp was selected and executed by the robot.
Otherwise, the sampling distribution was updated using the top $M\%$ of elite candidates and the process was repeated until a grasp was found or a maximum number of iterations was reached. 
During evaluation, the best available candidate was selected if no grasp reached the success criteria.

\subsection{Action Sampler}
In general, grasp candidates are generated by sampling from an action-space that describes the task. For this study we defined a specific task-space appropriate for top-down grasping from a table. A grasp candidate was represented by a four dimensional vector $s=[x, y, \psi, \alpha]$, where $x$ and $y$ corresponded to the planar position of the object on the table, $\psi$ was the gripper yaw angle, and $\alpha$ was percentage of gripper closure where 0 was fully open and 1 fully closed. Bounds on the action-space were set by assuming information about the workspace, such as table height, width, and length, and denoted as $\mathcal{W}$.

Given a sampling distribution $\mathcal{D}$, we generated a batch of $N$ top-down grasp candidates. Samples that were outside the workspace $\mathcal{W}$ or kinematically unfeasible were filtered out leaving a subset of allowed samples, $\hat{N}$. This process was repeated until a desired batch size $N^*$ was obtained.

\subsection{Action Image Representation}
\label{sec:representation}
In this section we describe the core contribution of this paper. \textit{Action Image} representation combines the power of convolutional networks to capture the spatial relationship of a grasp hypothesis with the local appearance and shape of objects. This process is outlined in Fig.~\ref{fig:preprocessor}.
In general, any salient feature set that expresses relevant grasping data and is expressible in image-space, such as rendering of gripper wireframes or CAD models or simple points could be used by this approach. For this work we chose to encode information about the gripper using feature points (see Fig.~\ref{fig:representation_kinematics}).

For a given grasp candidate $s$, we can use the robot kinematics to determine the transform from the end-effector $\mathcal{E}$ to a feature point $\xi_k$, denoted as $T_{\xi_k}^\mathcal{E}(s)$, and the transform from the robot frame $R$ to the end-effector, denoted as $T_\mathcal{E}^R(s)$. Assuming the camera is calibrated to the robot frame, we can also use the robot kinematics to find the camera frame $C$ relative to the robot frame as $T_R^C$. Combining these, we can find the location of each feature point in the camera frame as:

\vspace{-2mm}
\begin{equation}
T_{\xi_k}^{C}(s) = T_R^C T_\mathcal{E}^R(s) T_{\xi_k}^\mathcal{E}(s)
\vspace{-1mm}
\end{equation}

These feature points can in turn be projected onto the camera image plane through the camera intrinsic projection matrix. As a result, the \textit{Action Image} feature points can be synthesized for any grasp candidate without the need to execute the grasp. 

In this study, we described our specific pinch gripper using 3 feature points as can be seen in Fig.~\ref{fig:representation_kinematics}. Only 3 points were selected to provide the simplest graphical representation sufficient to infer the top-down grasp pose and opening of the gripper.
 
In addition to the spatial features of an \textit{Action Image}, the value of the rendered feature pixels can embed additional information relevant to the sensory modality. In our work, the feature points were rendered as a simple binary mask channel for the models with RGB information alone. In this case depth information was implicit, encoded only through the projection of the 3D points. In contrast, the D and RGB-D models directly set depth values for the feature patches and explicitly encoded this information. This showcases another advantage of the 
\textit{Action Image} representation, since it allows the network to exploit the intrinsic structural assumptions of convolutional neural networks and supports encoding prior knowledge in the value of the pixels.

The final piece of the representation used in this study was to exploit the locality of the image-based sensory information (RGB, D, RGB-D) by cropping image-data and \textit{Action Image} tightly around the \textit{gripper features}. 
Throughout the paper, we chose a crop size of $128 \times 128$ pixels as it provided a good compromise between including enough local information to evaluate a grasp candidate while avoiding unnecessary details about objects further out.
In practice, the application of CEM-based grasp optimization to our naive input cropping approach was sufficiently fast, resulting in inference times on the order of a few milliseconds. 

\subsection{Critic Network Architectures}
The architecture employed in this study was a composable critic network that allows the assembly of feature towers to rapidly incorporate different sensory modality inputs. Fig.~\ref{fig:nets}) illustrates the architecture for the three \textit{Action Image} networks proposed in this work and the architecture for the baseline following the design proposed in \cite{levine2018learning}.

Each \textit{Action Image} network was composed of a collection of feature towers from the set of: 1) image tower, 2) depth tower, 3) RGB action image tower, 4) depth action image tower, and 5) post-merge tower, Fig.~\ref{fig:towers} shows the architectural details.
These towers used convolution or ResNet layers as their building blocks (see Fig.~\ref{fig:resnet}). The output from the first 4 feature towers was a $32 \times 32 \times 32$ tensor. The post-merge tower was the common tower across all networks. This tower took a $32 \times 32 \times 32$ tensor input and produced a grasp success probability.

The baseline model, called action-vector, differed from the \textit{Action Image} networks, since it represented a grasp as a vector of floats (see Fig.~\ref{fig:action_vector_net}). In this network, action-vector feature tower consisted of the input, a grasp vector, passing through a series of fully connected layers (size 256 and 32) and then concatenating with the output from the image tower by repeating the same feature vector across the 32x32 spatial representation.

\section{Model Training}
\label{sec:model_training}

\begin{figure}[t]   
  \begin{center}
  \includegraphics[width=\linewidth]{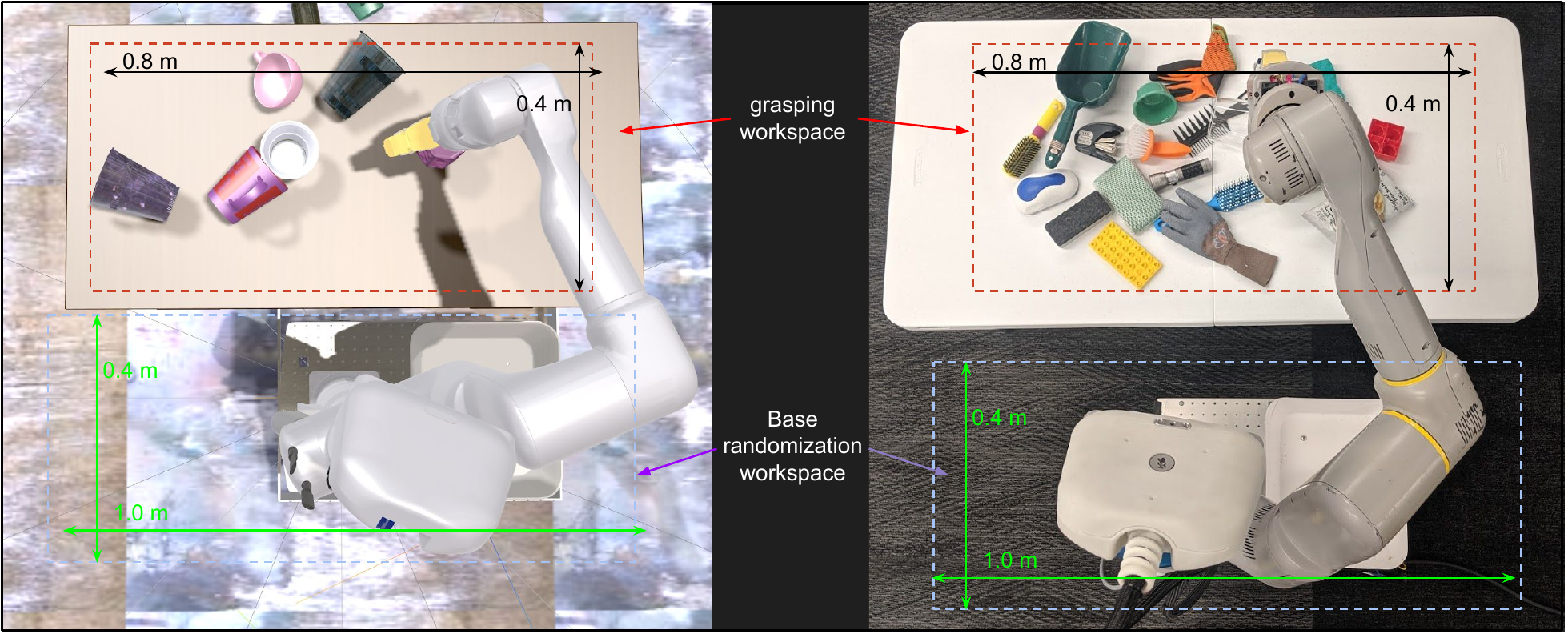}
  \end{center}
  \caption{\footnotesize{Workspace and random base footprints for simulated (left) and real-world (right) scenarios. \vspace{-0.3cm}}}
  \label{fig:workspace}   
\end{figure}

\begin{figure}[t]   
  \begin{center}
  \includegraphics[width=0.8\linewidth, trim=0 0cm 0 1.3cm, clip]{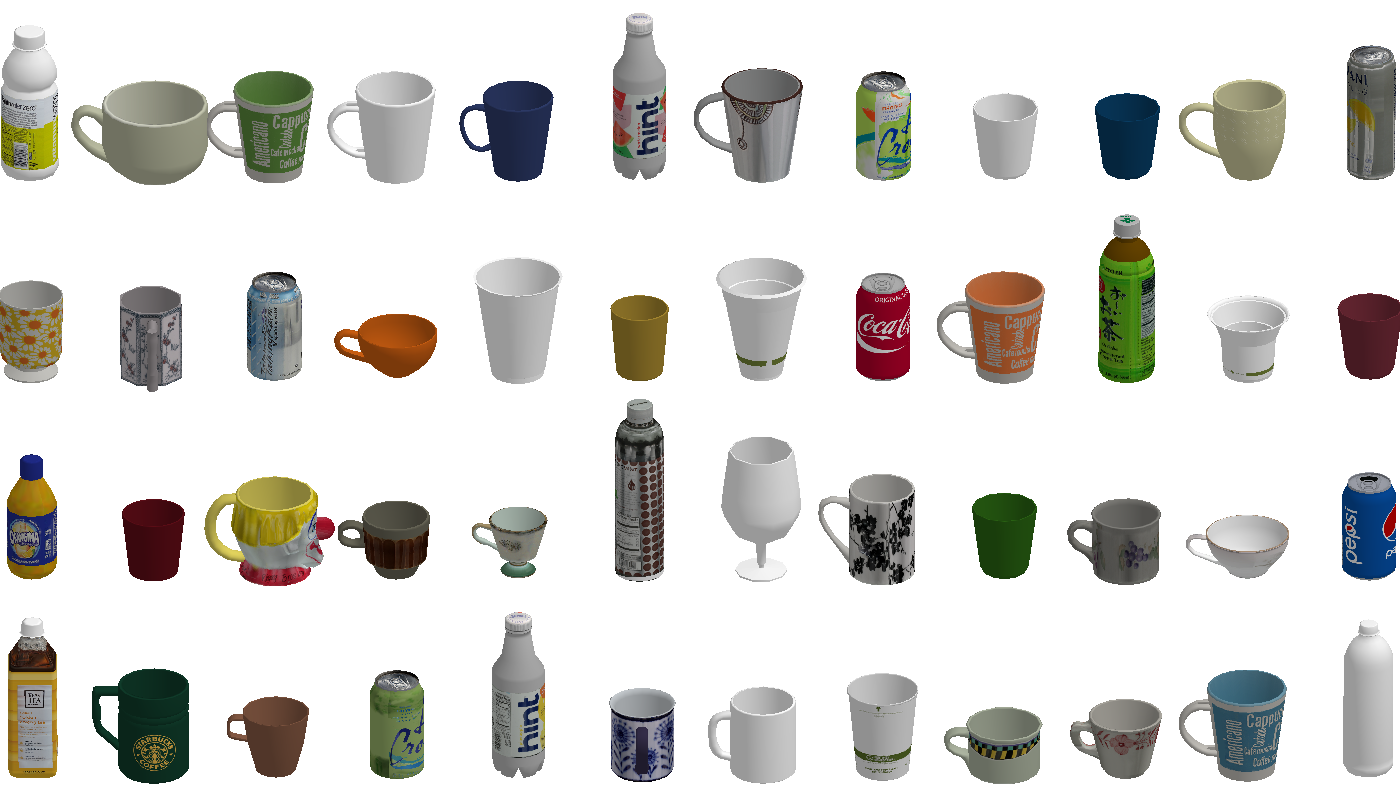}
  \end{center}
  \caption{\footnotesize{Set of 48 simulated objects used in the training of all models. \vspace{-0.3cm}}}
  \label{fig:simulation_objects}   
\end{figure}

Supervised training of the critic networks was accomplished using 3 datasets generated \textit{solely in simulation}.
No real world data was used during training for any of the models.
We used Bullet to simulate the task \cite{coumans2016pybullet}. Fig.~\ref{fig:workspace} shows the simulation setup for the data collection in sim.
Fig.~\ref{fig:simulation_objects} illustrates our training object dataset consisted of $48$ objects distributed between four object classes: bottles (8), cans (6), cups (19), and mugs (15).
For each run, the robot moved to a random base location within its base workspace illustrated in Fig.~\ref{fig:workspace}.
From the set of objects, a random number was drawn from the range of $[8, 12]$ and distributed randomly within the workspace.
Objects were initialized slightly above the table and allowed to fall, introducing additional randomness for their position and orientation. This resulted in some objects being placed outside the workspace or on the floor.
Domain randomization was applied to the table, floor and objects by shuffling the texture and color during each run.

\begin{algorithm}[tb]
  \caption{\footnotesize Scripted data collection policy in sim.}
  \label{alg:scripted}
  \scriptsize
  \begin{algorithmic}[1]
  \Require $\epsilon$, $\sigma_x$, $\sigma_y$, $\sigma_\psi$, $\sigma_\alpha$ and workspace $\mathcal{W}$
  \State Create a new scene with $n \sim \text{Uniform}(8, 12)$ objects.
  \State Store a snapshot of the scene $\mathcal{I}$.
  \State Draw $p \sim [0, 1]$
  \If{$p>\epsilon$}:
      \State Get objects position: $[x_i, y_i, \psi_i], i \in [0, n]$.
      \State For each object generate grasp pose by:\newline
        $\textrm{~~~~~~~~~} \beta_i = [\mathcal{N}(x_i, \sigma_x),$ $\mathcal{N}(y_i, \sigma_y),$ $\mathcal{N}(\psi_i, \sigma_\psi)]$.
      \State Randomly select a Kinematically feasible grasp pose\newline
        $\textrm{~~~~~~~~~} \beta \in \{\beta_0, \cdots, \beta_n\}$.
  \Else
      \State Draw a feasible top-down pose from the workspace.
  \EndIf
  \State Generate gripper opening angle $\alpha = \mathcal{N}(0.5, \sigma_\alpha)$,
  \State Set the grasp pose $s \leftarrow [\beta, \alpha]$
  \State Execute the grasp pose $s$.
  \State Determine the grasp success $l_s$ by lifting the object.
  \State {\bfseries Return} $\mathcal{I}$, $s$ and $l_s$
  \end{algorithmic}
\end{algorithm}

Curriculum training was employed for each model to generate a rich collection of $7.2$ million samples containing both grasp success and failures (Table \ref{tab:dataset}).
The initial bootstrap dataset (off-policy) contained $2.7$ million training examples collected in simulation using a scripted grasping policy using Algorithm \ref{alg:scripted}.
We trained all of the networks by using an ADAM optimizer on the cross-entropy loss between the estimated outcome from $\mathcal{G}$ and the true label $\mathcal{O}$. We used batch size 32 and balanced each batch with equal number of successful and failed grasp data. Photometric image distortion was also applied to both RGB and depth images.

Next, each trained network was used to generate on-policy datasets according to Algorithm \ref{alg:policy}. Please refer to Fig.~\ref{fig:architecture} and Section~\ref{sec:method} for further details on how a grasp candidate was selected.
By pooling the $850$ thousand grasps from each of the 4 models the on-policy \#1 dataset of $3.4$ million grasps was created.
Finally, each network was trained on both the bootstrap and on-policy \#1 datasets and in a similar fashion the on-policy \#2 dataset of $1.1$ million grasps was generated.

\begin{table}[tb]
\setlength{\tabcolsep}{0.3em}
\centering
\vspace{-0.1cm}
\caption{\footnotesize A list of simulation datasets and their sizes. On-policy data \#1 and \#2 were collected from policies trained on Dataset A and B, respectively. \vspace{0.1cm}}
\label{tab:dataset}
\footnotesize{
    \begin{tabular}{l|c|c|c|c}
                  & scripted policy & on-policy \#1 & on-policy \#2 & total \# grasps \\
    \hline\hline
    Dataset A   & 2.7m            &    -         & -      & \textbf{2.7m}       \\
    Dataset B   & 2.7m            & 3.4m         & -      & \textbf{6.1m}       \\
    Dataset C   & 2.7m            & 3.4m         & 1.1m   & \textbf{7.2m}       \\
    \end{tabular}
}
\vspace{-0.1cm}
\end{table}

\begin{algorithm}[tb]
  \caption{\footnotesize On-policy data collection and eval in sim.}
  \label{alg:policy}
  \scriptsize
  \begin{algorithmic}[1]
  \Require Policy $\mathcal{G}$, initial sampling distribution $\mathcal{D}, and $ workspace $\mathcal{W}$
  \State Create a new scene with $n \sim \text{Uniform}(8, 12)$ objects.
  \State Store a snapshot of the scene $\mathcal{I}$.
  \State Get the grasp pose $s \leftarrow \mathcal{G}(\mathcal{I};\mathcal{D}, \mathcal{W})$.
  \State Execute the grasp pose $s$.
  \State Determine the grasp success $l_s$ by lifting the object.
  \State {\bfseries Return} $\mathcal{I}$, $s$ and $l_s$
  \end{algorithmic}
\end{algorithm}
\section{Experiments And Results} \label{sec:experiments}
We developed a set of experiments to explore whether an \textit{Action Image} representation can improve performance over a direct-abstract representation and tested its performance in domain transfer from simulated to real-world scenes. 
Simulation evaluation was performed for the sole purpose of comparing the best possible performance across the different models. 
Testing on-robot was used to evaluate both domain transfer to real-world data and generalization to new objects.
Furthermore, we compared 3 \textit{Action Image} representations to evaluate the efficacy of the method to fuse different image information. 
We also evaluated against purely random and action-vector baselines to demonstrate the relative performance of the method.

\subsection{Simulation Evaluation}
The simulation evaluation was performed according to Algorithm \ref{alg:policy}. 
No hold-out objects and no real-world data were used during simulation evaluation.
The same set of objects used during training were used, but due to randomization of the color, texture, and pose, each scene was unique and different from those at training time.

Table~\ref{tab:sim_eval} provides a summary of the evaluation in simulation.
All three \textit{Action Image} models reached $+75\%$ success rate when trained only from scripted policy data.
Using on-policy data improved all models to $+85\%$ success rate and action-depth network had the best performance at $89\%$. 
As hypothesized, the baseline model was less data-efficient and only achieved $20\%$ success rate using the scripted policy data.
The baseline model performance improved significantly ($69\%$) when trained on all data, i.e. Dataset C, but did not reach the success rates of Action Image models that had been trained on much less data. This indicated the superior capability of Action Image representation in extracting grasp-relevant features compared to the baseline.

\begin{table}[tb]
\setlength{\tabcolsep}{0.3em}
\centering
\caption{\footnotesize Evaluation results in simulation. Success rate for each model is based on +5K grasp evaluations.}
\footnotesize{
    \begin{tabular}{l|c|c|c}
    & Dataset A & Dataset B & Dataset C \\
    \hline\hline
    Action RGB image         & \textbf{79.09\%} & 85.08\%          & 85.47\%          \\
    Action depth image       & 75.04\%          & \textbf{87.42\%} & \textbf{89.27\%} \\
    Action RGB-D image       & 76.38\%          & 85.32\%          & 87.85\%          \\
    Action Vector (baseline) & 20.45\%          & 60.53\%          & 69.06\%          \\
    \end{tabular}
\label{tab:sim_eval}
}
\end{table}

\subsection{On-robot Real-world Evaluation} \label{sec:real_eval}
\begin{figure}[t]   
  \begin{center}
  \begin{subfigure}[b]{0.45\linewidth}
    \includegraphics[width=\linewidth]{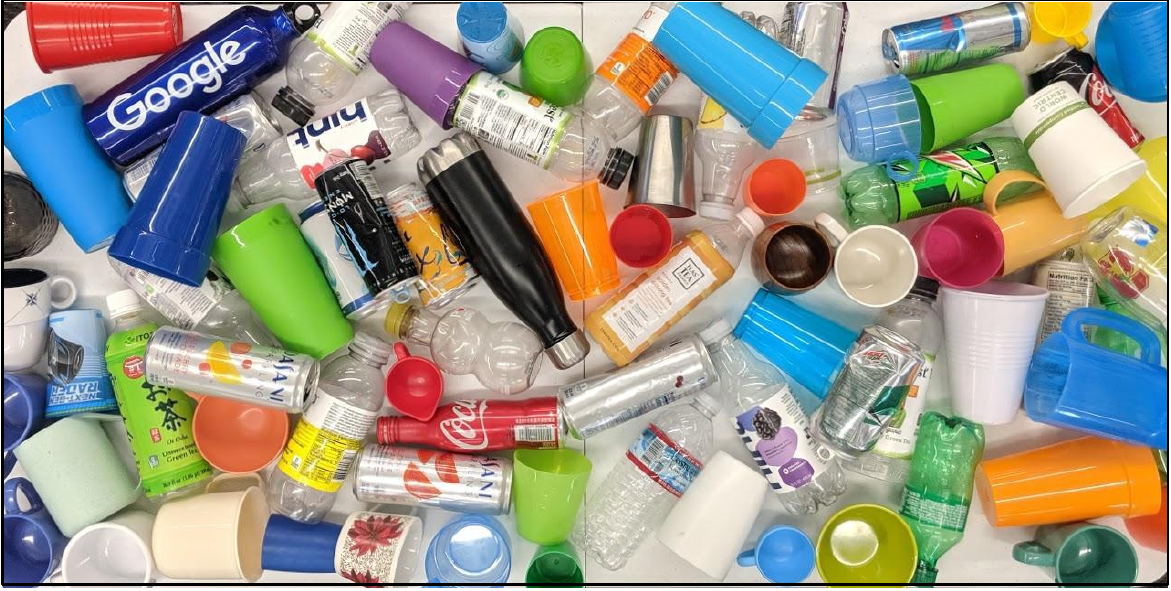}
  \end{subfigure}
  \begin{subfigure}[b]{0.45\linewidth}
    \includegraphics[width=\linewidth]{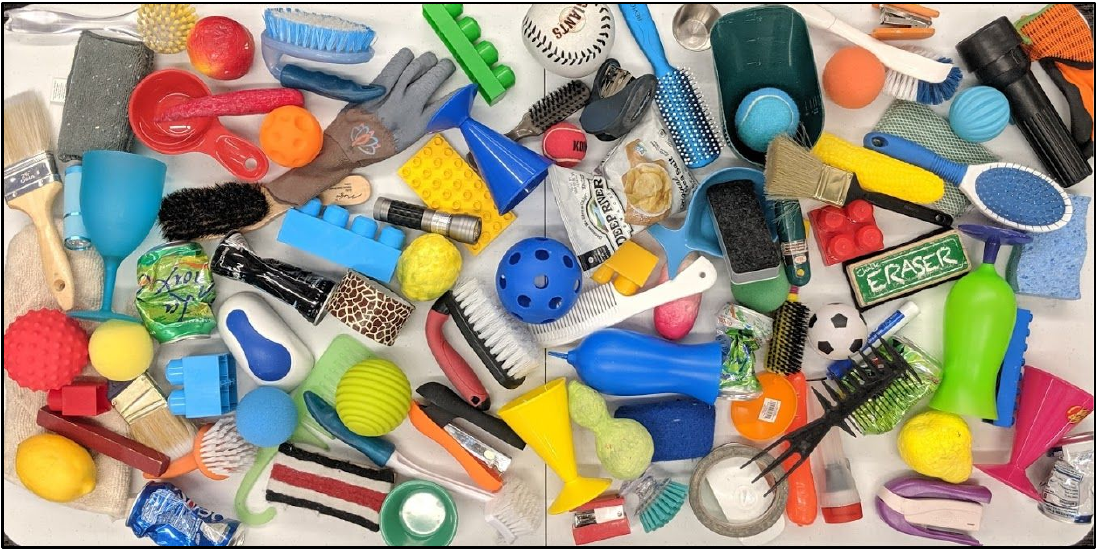}
  \end{subfigure}
  \end{center}
  \caption{\footnotesize{In total 172 objects are used in the real robot evaluation, including 80 familiar (left) and 92 novel (right) objects.}}
  \vspace{-4mm}
  \label{fig:real_world_objects}   
\end{figure}

The real-world experiments were performed using a cluster of 8 mobile manipulators, see Fig. \ref{fig:af_experiment}.
On-robot evaluation was similar to the simulated data collection setup.
First, the robot moved to a random pose in front of the table and then executed the grasping policy (see Fig. \ref{fig:workspace}).
To evaluate the generalization capabilities of the different policies, we considered two object datasets (see Fig. \ref{fig:real_world_objects}): 
1) A \textit{familiar} set that included 80 objects from the classes of bottles, cans, cups, and mugs. These objects were similar, but not identical to the simulation dataset.
2) A \textit{novel} set that included 92 random house-hold objects ranging from staplers, gloves, and sponges to combs, flashlights, and crushed cans. 
Both familiar and novel datasets had different visual properties from the simulated data, since the simulation did not render transparency or high reflectivity.
The novel dataset differed significantly from the simulation dataset in all aspects.
It contained deformable objects, gloves; thin objects, combs; and complex concave objects, scoops.

\begin{figure}[t]   
  \begin{center}
  \includegraphics[width=0.8\linewidth]{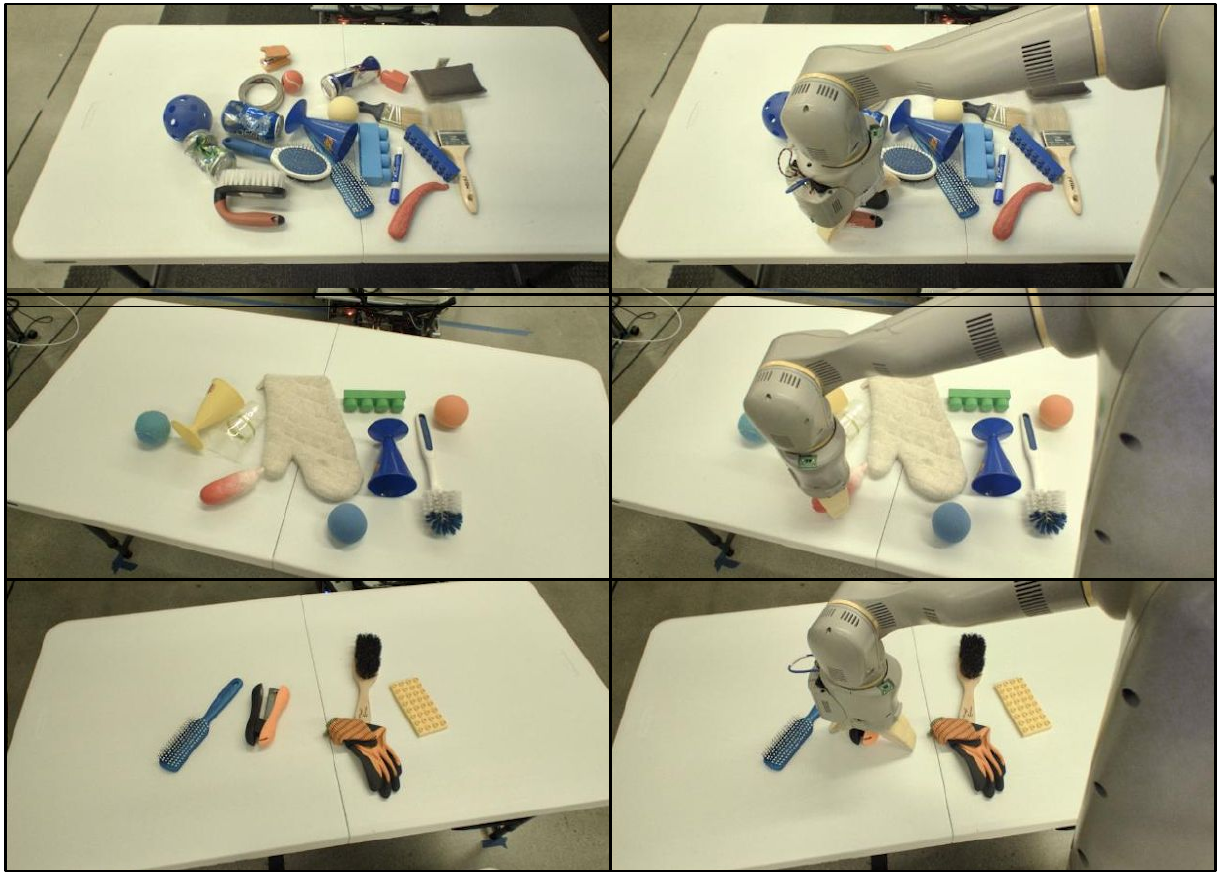} 
  \end{center}
  \caption{\footnotesize{Snapshots of the grasping attempts on novel objects in the scenes large, medium, and small (from top to bottom). Images correspond to the initial scene (left) and the time of grasp (right). \vspace{-0.3cm}}}
  \label{fig:af_experiment_hard}   
\end{figure}

We considered three different scenario sizes each with an increasing number of objects, small (5), medium (10) and large (20), see Fig.~\ref{fig:af_experiment_hard}.
For each evaluation run, we initialized a scene with one of a small, medium or large number of objects and drawn from either a familiar (objects similar to the training set) or novel (objects unlike those seen during training) set.
This produced 6 different experimental setups: 3 levels of object count $\times$ 2 object sets.

After each scene setup, the robot made 3 (small), 5 (medium), or 10 (large) grasp attempts based on the size of the scene.
Successfully grasped objects were placed in a basket behind the robot. 
Only objects placed in the basket were counted as success.
Objects that fell out of the gripper during placing in the basket were counted as failure.
Objects that were disturbed and fell off the table during grasp attempts were not returned to the table.
This implicitly penalized grasp mistakes by reducing the available objects.

In total, we performed 18 \textit{Action Image} evaluations, 3 scene sizes $\times$ 2 object sets $\times$ 3 networks, and an additional 12 baseline evaluations, 3 scene sizes $\times$ 2 object sets $\times$ 2 benchmark policies (action-vector and pure-random). 
The random policy was simply to query a single sample from the action sampler and execute the proposal.
Table \ref{tab:real_eval} reports the quantitative results of the real-world experimental evaluation based on 2160 grasps in real world.
The success rate $5.4\%$ from the random policy provides the lower-bond and illustrates the difficulty of the task and the experiment setup.
Overall, every \textit{Action Image} model achieved a success rate over $80\%$, while the action-vector baseline model only achieved a $53.4\%$ success rate.
Note, \textit{Action Image} performance transferred well from sim-to-real, with between $4-8$ percent points decrease whereas Action Vector dropped by $15$ percent points. 
This indicated that \textit{Action Image} generalized better than Action Vector in real world scenarios.

The results suggest that the medium-sized scene with familiar objects correspond to the easiest scenario. This is likely because familiar objects were generally easier to grasp and the medium-sized scenario provided a goldilocks case where there was a minimum of clutter and plenty of objects to grasp. 
The small scene was more difficult than the medium scene due to the scarcity of the objects. The base randomization at the beginning of each grasp could potentially make some of the objects kinematically unreachable, hence giving the network less options for grasping.
As expected the large scene with novel objects correspond to the hardest scenario due to the difficulty finding a good grasp for an object and the significant visual and physical differences between the novel and familiar objects.

Interestingly, all models, including action-vector, generalized well to the novel objects.
However, \textit{Action Image} models with depth data were relatively more over-fit.
In the medium-sized scene action-RGB-D decreased in performance for the novel objects (familiar at 95\% vs. novel at 86\%) while action-RGB actually improved slightly (familiar at 81\% vs. novel at 84\%).

\begin{table}[tb]
\setlength{\tabcolsep}{0.3em}
\centering
\vspace{-0.5cm}
\caption{\footnotesize Quantitative evaluation of on-robot testing. These results were based on 2160 grasps in total. The success rate for each scenario was calculated from, on average, $83$ grasps, except for the random policy where only, on average, $28$ grasps were performed.}
\scalebox{0.9}{
    \begin{tabular}{l|c|c|c|c|c|c||c}
       & \multicolumn{3}{c}{familiar} & \multicolumn{3}{|c||}{novel} & \\
       & small & medium & large & small & medium & large & Overall\\
    \hline\hline
    \multirow{2}{*}{Action RGB}    & {85.3\%}        &        {81.1\%} & \textbf{88.8\%} & {74.2\%}        &        {83.5\%} & {71.3\%}        & \multirow{2}{*}{80.6\%} \\\cline{2-7}
                                   & \multicolumn{3}{c}{85.1\%} & \multicolumn{3}{|c||}{76.2\%} & \\
    \hline\hline
    \multirow{2}{*}{Action depth}  & {83.9\%}        &        {91.7\%} &        {82.7\%} & \textbf{81.5\%} &        {83.1\%} & {64.4\%}        & \multirow{2}{*}{81.3\%} \\\cline{2-7}
                                   & \multicolumn{3}{c}{85.8\%} & \multicolumn{3}{|c||}{76.9\%} & \\
    \hline\hline
    \multirow{2}{*}{Action RGB-D}  & \textbf{91.6\%} & \textbf{94.8\%} &        {83.3\%} & {76.1\%}        & \textbf{86.3\%} & \textbf{74.4\%} & \multirow{2}{*}{\textbf{84.4\%}}\\\cline{2-7}
                                   & \multicolumn{3}{c}{\textbf{90.0\%}} & \multicolumn{3}{|c||}{\textbf{78.8\%}} & \\
    \hline\hline
    Baseline:                      & {62.1\%}      &        {51.9\%} &        {58.2\%} & {40.2\%}        &        {53.8\%} & {54.5\%}        & \multirow{2}{*}{53.4\%}\\\cline{2-7}
    Action Vector                  & \multicolumn{3}{c}{57.8\%} & \multicolumn{3}{|c||}{49.0\%} & \\
    \hline\hline
    \multirow{2}{*}{Baseline: Random}        & {4.2\%}         &        {15\%}    &        {7.5\%} & {0\%}           &        {0\%}    & {5.0\%}         & \multirow{2}{*}{5.4\%}\\\cline{2-7}
                                   & \multicolumn{3}{c}{8.3\%} & \multicolumn{3}{|c||}{2.4\%} & \\
    \hline\hline    
    \end{tabular}
}
\label{tab:real_eval}
\end{table}

\section{Conclusion} \label{sec:conclusion}
In this paper we hypothesized that the action representation for visual-based grasping is key to learning critic networks efficiently and allowing them to generalize.
We introduced \textit{Action Image}, a visual representation of grasps to better exploit the structural assumptions implicitly encoded in convolutional neural networks.
By simply rendering gripper feature into image-space and providing a cropped, focused image-based information to the network we achieved over $84\%$ real-world grasping performance without training on any real-world examples.
In our experiments, we showed that \textit{Action Image} representations can easily incorporate additional image-based information, such as depth, and outperforms vector-based action representations.
We confirmed an established observation that grasping is an object-centric, local problem by demonstrating that \textit{Action Image} with image-space cropping and diverse simulation training data can achieve high rates of grasp success. Furthermore, we provided more evidence that domain transfer and object generalization is achievable, strengthening the theory that grasping behaviors can be learned in simulation and transferred to real-world robots.

\bibliographystyle{ieee}
\bibliography{main}  

\end{document}